\RequirePackage{fix-cm}
\documentclass[smallextended]{svjour3}       
\smartqed  
\usepackage{graphicx}
\usepackage{algorithmic}
\usepackage{algorithm}
\usepackage{amssymb}
\usepackage{color}
\usepackage{array}
\usepackage{epstopdf}
\usepackage{subfigure}
\usepackage{algorithm}
\usepackage{algorithmic}
\usepackage{multirow}
\usepackage{amsmath}
\usepackage{hyperref}

\newcommand{\eat}[1]{}

\begin{document}

\title{Exploiting Long-term Temporal Dynamics for Video Captioning 
}


\author{Yuyu Guo       \and Jingqiu Zhang \and
        Lianli Gao 
}


\institute{	
           Lianli Gao \at     
          \email{lianli.gao@uestc.edu.cn}    
          \\
          \\
           Yuyu Guo,Jingqiu Zhang and Lianli Gao are with the Future Media Center and School of Computer Science and Engineering, The University of Electronic Science and Technology of China. \\
          Lianli Gao is the corresponding author     
}

\date{Received: date / Accepted: date}

\maketitle

\begin{abstract}
Automatically describing videos with natural language is a fundamental challenge for computer vision and natural language processing. Recently, progress in this problem has been achieved through two steps: 1) employing 2-D and/or 3-D Convolutional Neural Networks (CNNs) (e.g. VGG, ResNet or C3D) to extract spatial and/or temporal features to encode video contents; and 2) applying Recurrent Neural Networks (RNNs) to generate sentences to describe events in videos. Temporal attention-based model has gained much progress by considering the importance of each video frame. However, for a long video, especially for a video which consists of a set of sub-events, we should discover and leverage the importance of each sub-shot instead of each frame. In this paper, we propose a novel approach, namely temporal and spatial LSTM (TS-LSTM), which systematically exploits spatial and temporal dynamics within video sequences. In TS-LSTM, a temporal pooling LSTM (TP-LSTM) is designed to incorporate both spatial and temporal information to extract long-term temporal dynamics within video sub-shots; and a stacked LSTM is introduced to generate a list of words to describe the video. Experimental results obtained in two public video captioning benchmarks indicate that our TS-LSTM outperforms the state-of-the-art methods.

\eat{
Recently, the video captioning problem has attracted more and more attention and automatically generating video description with natural language is a fundamental task of computer vision. 
Using deep neural networks is a powerful way to solve the video captioning problem. And most of existing works use pretrained deep convolutional neural networks (e.g. VGG, ResNet and C3D) to extract spatial features and temporal features, then use RNNs to generate the words.

 Recently, the temporal attention model, which concerns about the differences between frames and frames, has achieved great results for solving the videos captioning problem. But for a short video clip, it may not contain  many temporal variances and few people explore the temporal invariance on video captioning. In this paper, we design a effective framework based on deep neural networks for video captioning and the architecture can maintain the temporal invariance and variance. And combining effective temporal features, our model achieves better results on video captioning benchmarks.
}
\end{abstract}

\section{Introduction}
\label{sec:intro}
With the development of multimedia and information technology, huge amounts of videos are uploaded and downloaded on the Internet every day, thus
it has spawned a great deal of research into videos or images, such as video/image classification \cite{CLA:zhu}, video/image retrieval \cite{Hashing:zhu2013linear,Hasing:songhe,Hasing:GF_song,Hasing:zhu2014sparse}, video segmentation \cite{Seg:song_L,Seg:Gao_graph}, video annotation \cite{Anno:Gao_song,Anno:Song_gao} and video captioning \cite{CAP:TMM_gao,CAP:adpt_Song} etc.  As a bridge between computer vision and natural language, video captioning has become a hot research topic in recent years. Moreover, describing video contents with natural language becomes a key component for improving human-robot interaction and artificial intelligence. To date, extracting features from videos and then translating them into natural language sentences is the main trend, therefore researchers \cite{VCAP_Trand:Khan2011Human,VCAP_Trand:Lee2008SAVE,VCAP_Trand:Thonnat,VCAP_Trand:Kojima2002} are focusing on solving two sub-problems: 1) how to efficiently extract video features; and 2) how to accurately translate video features into sentences with Recurrent Neural Networks (RNNs).

\begin{figure}
	\centering
	\includegraphics[width=0.9\linewidth]{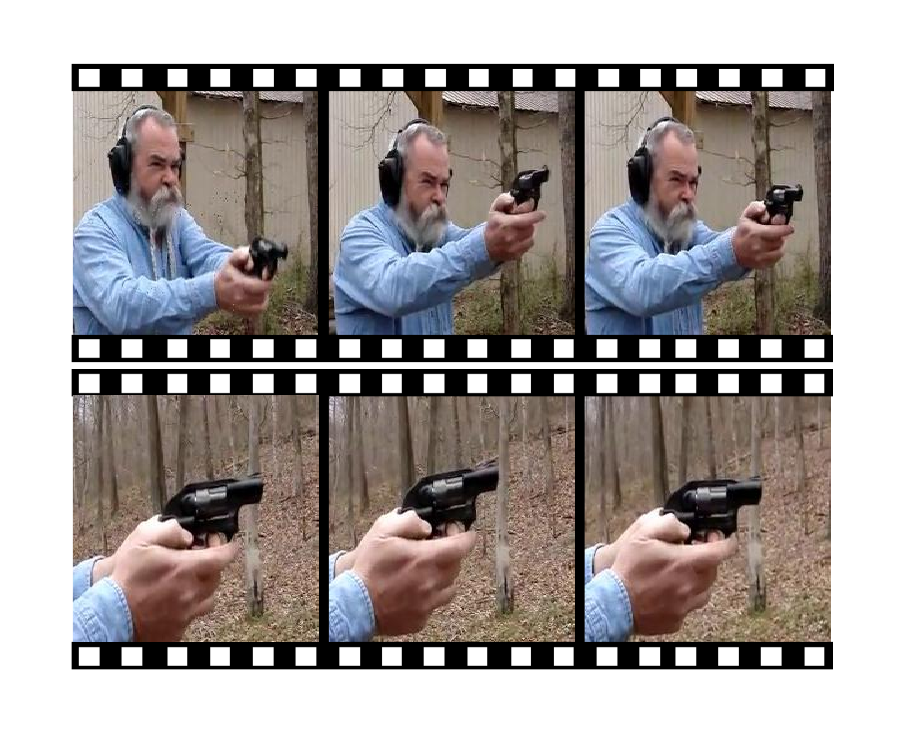}
	\caption{We extract six frames from a video, which contains distinct appearance information. However, the top three frames do not contain much variance, so as the bottom three frames.}
	\label{Fig.samples}
\end{figure}

Specifically, \cite{VCAP_Trand:Khan2011Human,VCAP_Trand:Thonnat} firstly identified video semantic contents and then generated sentences based on some templates. \cite{VCAP_Trand:Lee2008SAVE,VCAP_Trand:Farhadi_MRF,VCAP_Trand:Rohrbach_CRF} disposed the problem with probabilistic graphical model (PGM) and utilized Markov random field  (MRF) or Conditional random field (CRF) to find the relationship between visual content and natural language. Great success in image classification achieved by deep convolutional neural networks (e.g. GoogLeNet  \cite{Net:GNetSzegedy_2015_CVPR}, VGG \cite{Net:VGGSimonyan2014Very} and ResNet \cite{Net:ResNetHe_2016_CVPR}), these networks provide researchers with powerful tools to extract image/video features on different fields, such as video captioning \cite{CAP:EmbedPan2015Jointly,CAP:HRNNPan2016Hierarchical} and video action recognition \cite{VACT:TS-LSTM,VACT:TSN}. In general, the basic video captioning framework adopts pre-trained deep CNNs (e.g. ResNet or C3D) to extract spatial and/or temporal features, and then applies an RNN network (e.g. LSTM\cite{RNNs:LSTMsep}, GRU\cite{RNNs:GRUJun} or their extensions.) to generate words.

Furthermore,  Mnih \textit{et al.} \cite{Att:RNN_VA}  proposed a novel recurrent neural network model to extract information from images by adaptively selecting a sequence of regions or locations and only processing the selected regions at high resolution. The experimental results showed that it significantly outperforms the convolutional neural network baseline on a dynamic visual control problem. This strategy is named as visual attention. In fact, the basic idea of the attention mechanism is to selectively focus on the important information and  maximumly ignore the unimportant information in the meantime. Therefore, the first step of attention is to estimate which part is important and assign a higher weight to it. Inspired by its great success, a variety of visual attention models have been proposed \cite{CAP:adpt_Song,CAP:softattYao2015Describing}. For example, \cite{CAP:adpt_Song,CAP:softattYao2015Describing} introduced a temporal attention to enhance video captioning by setting frames with different weights to select the most relevant temporal segments by training. However, for a video, the temporal variances are existing between sub-shots instead of adjacent frames. From Fig. \ref{Fig.samples}, we can see that all of the frames are important for describing the video, but many of the frames are duplicate. The top three frames describe a man who is holding a gun, while the bottom three frames describe the shooting action. Weighting each frame would incur excessive computational cost and result in low accuracy. 

Here, we argue that long-range temporal structure plays an important role in understanding dynamics in video captioning. However, mainstream  video captioning frameworks  \cite{CAP:adpt_Song,CAP:softattYao2015Describing} usually focus on appearances and short-term motions, which lack the capacity to incorporate long-range temporal structure. In this paper, we aim to study the following problem: How to design an effective and efficient video-level framework for learning video representation that is able to capture long-range temporal structure for improving video captioning. Moreover, in terms of long-range temporal structure modeling, a key observation is that consecutive frames are highly invariant \cite{VACT:TSN,VACT:TS-LSTM} (Fig. \ref{Fig.samples}), thus it is unnecessary to directly set dense temporal sampling for LSTMs.  Therefore, we propose a novel framework, namely temporal and spatial LSTM (TS-LSTM), which firstly uses a temporal pooling (TP) layer to keep the temporal invariance in a short video shot, then a Long Short-Term Memory (LSTM) \cite{RNNs:LSTMsep} to exploit temporal dynamics between long-range video shots. In addition, a stacked Long Short-Term Memory (Stack-LSTM) is adopted to generate words in the final stage.  This framework employs representations from spatial and temporal features to enhance video captioning.  The contributions of this paper are as follows:
\begin{itemize}
	\item Given spatial and motion feature representations over time, we propose to integrate a temporal pooling and a LSTM to learn both temporal invariance and variance. This mechanism fuses high-level spatial and temporal features to learn long-range temporal dynamics over the whole video.  
	\item We introduce a TS-LSTM video captioning framework, which integrates TP-LSTM with a mean pooling and a stacked LSTM to automatically generate words for describing a video. Specifically, the mean pooling is applied on a concatenation of visual features, motion features and long-term dynamics to extract useful information for the decode process. In addition, inspired by the two-stream framework \cite{VACT:two-stream,VACT:two-streamfusion} which has achieved great results in video action recognition, we adopt a fine-tuned Resnet-152 \cite{Net:ResNetHe_2016_CVPR} to extract the temporal features. Compared with C3D \cite{Net:C3DDuTran} features, using Resnet-152 features can achieve better results.   
	\item We perform experiments on two video captioning datasets, namely MSVD \cite{Dataset:msvd} and MSR-VTT \cite{Dataset:msr-vtt}, to verify the effectiveness of our method. The experimental results show that our method outperforms existing approaches. 
\end{itemize}


\section{Related Work}

\label{sec:relate_work}
\subsection{Deep Convolutional Neural Network}
In the field of deep learning, deep convolutional neural networks (CNNs) have been widely applied to explore visual information, such as image recognition \cite{Net:LeNet}, object detection \cite{Nets:faster-rcnn} and image retrieval \cite{Hasing:songhe} etc. From LeNet \cite{Net:LeNet} to ResNet \cite{Net:ResNetHe_2016_CVPR}, the performances of such models have greatly improved on the task of image classification. Specifically, ResNet-152 \cite{Net:ResNetHe_2016_CVPR}  achieves better results than human beings. As a result, many researchers employ these networks to improve  the performance of their tasks. For example, Feichtenhofer \textit{et al.} \cite{VACT:two-stream} fine-tuned the VGG \cite{Net:VGGSimonyan2014Very} to improve the performance on video action recognition task, and Yao \textit{et al.} \cite{CAP:softattYao2015Describing} used a pre-trained GoogLeNet \cite{Net:GNetSzegedy_2015_CVPR} to extract features for video captioning. Motivated by the previous works \cite{CAP:MFA_long,CAP:adpt_Song}, we use the ResNet-152 to extract features both for spatial and temporal information. In addition, all the above mentioned deep CNNs contain pooling layers, which are always used to reduce the spatial size and solve the over-fitting problem. Besides, Scherer \textit{et al.} \cite{DNN:Pooling_Scherer} showed that pooling layers have potential to obtain spatial invariance, thus we integrate a temporal pooling layer to explore the temporal invariance in a video short snip in this paper.

\subsection{Recurrent Neural Networks}
Compared with CNNs, Recurrent Neural Networks (RNNs)  are good at modeling sequential data, thus they have been widely utilized in natural language processing and achieved great success \cite{RNNs:simpleRNN_2,RNNs:simpleRNN_1}. 
At each time step, an RNN observes an element and updates its internal states. In the field of speech recognition, the RNN Language Model (RNNLM) \cite{RNNs:RNNLM_Tomas} models the output distribution by adding a softmax layer onto the hidden states. In order to learn the RNNLM model's parameters, it maximizes the log-likelihood by using the gradient descent method. 

However, the above mentioned RNNs are suffering from the ``long-term dependencies'' problem \cite{RNNs:long-termYoshua}. LSTM \cite{RNNs:LSTMsep} is designed for leaning long-term dependencies. It solves the ``long-term dependencies'' problem by adding some gates that explicitly allow the RNN to learn when to forget previous hidden states with ``forget gate'' and when to update hidden states given new inputs. Previous studies showed that LSTM is capable of modeling data sequences, especially for encoding sentences and video features. Therefore, in this paper, we choose LSTM as our basic component for video captioning. 

\subsection{Video Captioning}
As a bridge connecting computer vision and natural language processing, video captioning has attracted great attention in both areas. How to auto-generate descriptions of images or videos is an old topic in computer vision \cite{VCAP_Trand:Farhadi_MRF,VCAP_Trand:Kojima2002,VCAP_Trand:Lee2008SAVE}. For example,  Kojima \textit{et al.} \cite{VCAP_Trand:Kojima2002} firstly detected human postures, including head positions, head directions and hands positions, and then several predicts and objects are selected with domain knowledge. Finally, they filled these syntactic components into a case frame, and translated the case frames into sentences with some syntactic rules. In addition, same strategy is utilized to enhance other multimedia applications, such as \cite{VCAP_Trand:Khan2011Human,VCAP_Trand:Hanckmann2012Automated}.

Later on, some researchers tried to describe videos/images with probabilistic graphical model \cite{VCAP_Trand:Lee2008SAVE,VCAP_Trand:Farhadi_MRF,VCAP_Trand:Rohrbach_CRF}. For instance, Farhadi \textit{et al.} \cite{VCAP_Trand:Farhadi_MRF} constructed three spaces: image space, sentence space and meaning space. In order to find the relationship between images and the corresponding sentences, they projected both image and sentence spaces into a common space: the meaning space. Specifically, the meaning space was represented by a triplet indicating as $<$object, action, scene$>$. Mapping the image space to the meaning space was reduced to predicting the triplets from images, while mapping the sentence space into a meaning space was conducted by extracting triplets from sentences and then computing the similarity between two triplets. In addition, Rohrbach \textit{et al.} \cite{VCAP_Trand:Rohrbach_CRF}
explored the relationship between visual contents and semantic representations with Conditional Random Field (CRF). However, all of these methods are highly dependent on the templates of sentences, which is insufficient to model the richness of natural language.

Recently, inspired by the great success of deep learning, many researchers \cite{CAP:Guo2016Attention,CAP:EmbedPan2015Jointly,CAP:softattYao2015Describing,CAP:S2VT} applied deep neural networks to solve the video captioning problem. 
 Specifically, Venugopalan \textit{et al.} \cite{CAP:S2VT}
employed a stacked LSTM for generating good descriptions effectively. The first LSTM encodes the visual features from pre-trained CNNs and the second LSTM generates words.
Pan \textit{et al.} \cite{CAP:EmbedPan2015Jointly} leveraged the semantics, both from entire sentence and video content,  to learn a visual-semantic embedding model.
Some works \cite{CAP:Pan_2017_CVPR,CAP:MFA_long} showed that semantic attributes make a significant contribution to video captioning. Pan \textit{et al.} \cite{CAP:Pan_2017_CVPR} adopted the Multiple Instance Learning (MIL) to learn the semantic attributes from videos, then utilized the generated attributes to improve the performance of their models. 
Compared with mean pooling, \cite{CAP:adpt_Song,CAP:softattYao2015Describing,CAP:Yu_2017_CVPR} were interested in tackling video captioning with attention mechanisms. Yao \textit{et al.} \cite{CAP:softattYao2015Describing} introduced a temporal soft attention mechanism into video captioning to  automatically select the most relevant frames.
Yu \textit{et al.} \cite{CAP:Yu_2017_CVPR} introduced a supervised spatial attention mechanism to guide the model to learn the relevant spatial information for video captioning.
Different with above works, we are focusing on further extracting informative features for videos in terms of exploiting a long-range temporal structure.  

\section{The Proposed Approach}
\label{sec:approach}
In this section, we introduce our approach for video captioning. Firstly, we define the terms and notations. Next, we describe our proposed network. Finally, we introduce the loss function of our model.
\begin{figure*}
	\centering
	\includegraphics[width=1.0\textwidth,height=6cm]{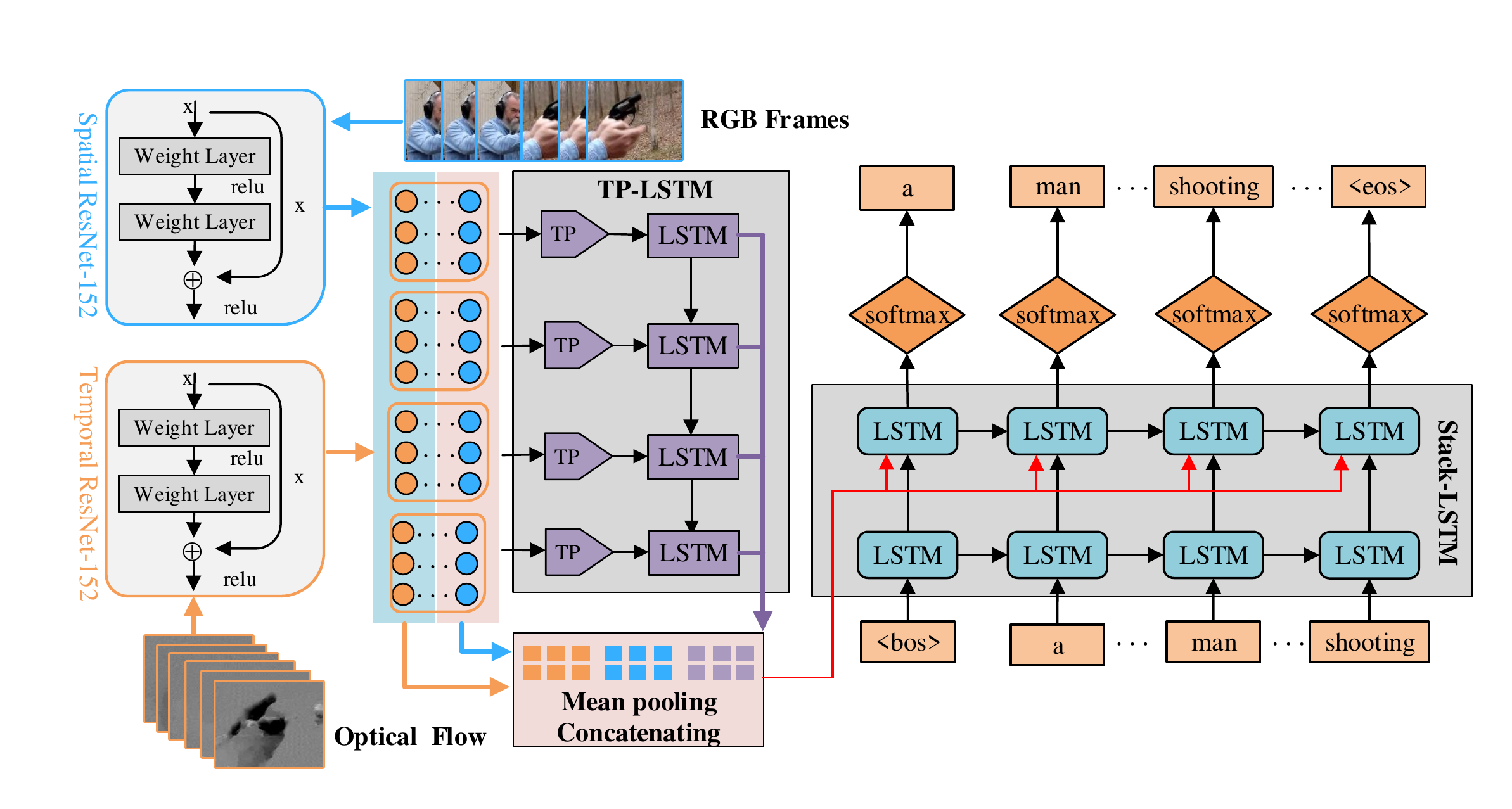}
	\caption{The framework of our model. TP-LSTM explores the invariance and variance in the video, while a Stack-LSTM is applied to generate words for describing the video. }
	\label{Fig.framework}
\end{figure*}

\subsection{Terms and Notations}
Given a video ${\bf{V}}$, we extract its features 
$ {\bf{V}} = \left\{ {{v_1}, {v_2} \cdots ,{v_i} , \cdots ,{{v}_{{N_v}}}} \right\} \in \mathbb{R}^{D_v \times N_v} $,
${D_v}$ denotes the dimension of visual features, $N_v$ denotes the number of sampled frames from the video. A sentence ${\bf{S}} = \left\{ {{s_1},{s_2}, \cdots , {s_i},\cdots ,{{s}_{{N_s}}}} \right\}  \in \mathbb{R}^{D_s \times N_s} $ consisting of ${N_s}$ words for describing the video, and $s_i$ is an one-hot vector. $D_s$  is the size of dictionary. 
\eat{At each time step, our model predicts one word based on generated feature and words that have been generated.}
And we denote $<$BOS$>$ as the start of a sentence. Our framework is shown in Fig. \ref{Fig.framework}. This framework consists of six major components. The first component is a Spatial ResNet-152 network which takes RGB frames as inputs and extracts visual features from each video frame, while the second component is the Temporal ResNet-152 network which takes optical flows as input and produces temporal features for each frame. Next, the third component is a concatenation that concatenates the outputs of  Spatial ResNet-152 and Temporal ResNet-152 networks. Then, TP-LSTM takes a set of concatenations as inputs with a temporal pooling strategy. Finally, the second concatenation integrates visual features, temporal features and the outputs of TP-LSTM into a new video representation. The last component is a stacked LSTM, which takes the new video representation and words to produce a natural language sentence.

\subsection{Temporal Pooling LSTM}
How to extract effective visual features is an important problem for analyzing videos. Due to the rapid development of deep convolutional neural networks (CNNs), which have made a great success in image classification \cite{Net:ResNetHe_2016_CVPR}, object detection \cite{Nets:faster-rcnn} and video action recognition \cite{VACT:two-stream}, it is common to apply deep CNNs to extract visual features. In this work, we use the ResNet-152 per-trained on the ImageNet to extract video frame visual features. In addition, a video contains not only spatial information but also temporal information. Therefore, we utilize another fine-tuned ResNet-152, which takes optical flow images as inputs, to extract video temporal features. After that, we concatenate above two features together. In order to model the invariance and variance of the input video, we propose a temporal pooling LSTM to dispose the fused new feature. More specifically, we divide the new features into $N_e$ parts along the temporal dimension, thus each part  has $N_k = N_v / N_e$ features. Next, we average the features from same part. This process is expressed as follow:
\begin{equation}
\begin{aligned}
e_i =  \frac {\sum_{j=(i-1)\times{N_k}}^{i \times {N_k}} v_j}{N_k} \ \ \ \ \ i \in \{1,2,...,N_e\} 
\end{aligned}
\end{equation}
${\bf{E}} = \left\{ {{e_1}, {e_2} \cdots ,{e_i} , \cdots ,{{e}_{N_e}}}\right\}  \in \mathbb{R}^{D_v \times N_e} $ is generated after the temporal pooling. 

In the next step, we aim to extract long-term dynamics across video by applying a Recurrent Neural Network (RNN) on ${\bf{E}}$.  As mentioned above, we employ LSTM to model the long-term temporal dynamics of ${\bf{E}}$. The structure of LSTM is described below:

\begin{equation}
\begin{array}{l}
f_t = \sigma(W_{xf} e_t + W_{hf} h_{t-1} + b_f) \\
i_t = \sigma(W_{xi} e_t + W_{hi} h_{t-1} + b_i) \\
o_t = \sigma(W_{xo} e_t + W_{ho} h_{t-1} + b_o) \\ 
g_t = \phi(W_{xg} e_t + W_{hg} h_{t-1} + b_g) \\
c_t = f_t \odot c_{t-1} + i_t \odot g_{t} \\
h_t = o_t \odot \phi \left( {{c_t}} \right)
\end{array}
\end{equation}
where $\sigma(\cdot) $ denotes the sigmoid function, $\phi(\cdot)$ denotes the hyperbolic tangent function, and $\odot$ denotes the element-wise multiplication. $c_t$ is a cell state vector, and $h_t$ is an hidden state vector.  $W_{*}$ is a set of parameters, and $b_*$ is a set of bias values.  For convenience, we define the function as:
\begin{equation}
\label{Eq.lstm}
\begin{array}{l}
 h_t, c_t = LSTM(e_t, h_{t-1}, c_{t-1};W,b) \ \ \ \ \ t \in \{1,...,N_e\} 
\end{array}
\end{equation} 
where $e_t$ is the input at $t$-th time step, and $h_0$, $c_0$ are initialized vectors.
In our model, we use $h_t$ as the output of the LSTM. After $N_e$ time steps, we get  ${\bf{H}} = \left\{ {{h_1}, {h_2} \cdots ,{h_i} , \cdots ,{{h}_{N_e}}}\right\}  \in \mathbb{R}^{D_h \times N_e} $.  $D_h$ is the output dimension of the LSTM. Next, we average the outputs of LSTM and visual features, respectively. See below:
\begin{equation}
\label{Eq.mean}
\begin{array}{l}
\overline{v} = \frac{\sum^{N_v}_{i=1} v_i}{N_v}   \ \ \ \ \ v_i \in \bf{V} \\
\overline{h} = \frac{\sum^{N_e}_{i=1} h_i}{N_e}   \ \ \ \ \ h_i \in \bf{H}
\end{array}
\end{equation}
Then, we concatenate them $y = [\overline{v},\overline{h}]$ and feed them into our Stack-LSTM.

\subsection{Stacked LSTM}
In order to reduce the dimension of the one-hot vector and explore the semantic information from the one-hot vector, we follow previous works \cite{CAP:Pan_2017_CVPR,CAP:p_RNNYu_2016_CVPR,CAP:adpt_Song} to embed the one-hot vector into a low-dimensional vector as follow:
\begin{equation}
\label{equ.embed}
\textbf{M} = {W_{s} \textbf{S}}
\end{equation}
where $W_{s} \in \mathbb{R}^{D_m \times D_s} $ is a parameter matrix. After embedding, we obtain an embedding matrix ${\bf{M}} = \left\{ {{m_1}, {m_2} \cdots ,{m_i} , \cdots ,{{m}_{N_s}}}\right\}  \in \mathbb{R}^{D_m \times N_s} $. 

Then we use LSTM layers to explore semantic information from both sentences and videos. Donahue \textit{et al.} \cite{IMCAP:Two_Donahue} suggested that two LSTM layers are better than one or four layers for image captioning. Compared with their two LSTM layers, our first LSTM layer is used to encode sentence information, while the second LSTM layer is applied to fuse both sentence  and visual information for achieving semantic features. More specifically, at first we use a standard LSTM to explore the relationship between words: 

\begin{equation}
\begin{array}{l}
 q_t, u_t = LSTM(m_t, q_{t-1}, u_{t-1};W_q,b_q) \ \ \ \ \ t \in \{1,...,N_s\} 
\end{array}
\end{equation} 
where $q_0$ and $u_0$ are initialized vectors. $W_q$ and $b_q$ are parameters. After $N_s$ time steps, we get a series of vectors  ${\bf{Q}} = \left\{ {{q_1}, {q_2} \cdots ,{q_i} , \cdots ,{{q}_{N_s}}}\right\}  \in \mathbb{R}^{D_q \times N_s} $, which contain temporal information from a sentence.
Next, we use a multi-modal LSTM (M-LSTM)  which incorporates features from different information sources (i.e., video and words) into a set of higher-level representations. The M-LSTM integrates information from visual and word sources into latent semantic features by adjusting their weights to improve the video captioning performance. The structure of the multi-modal LSTM is described as follows:
\begin{equation}
\begin{array}{l}
\label{euq.m-lstm}
f'_t = \sigma(W'_{xf} q_t + W'_{hf} h'_{t-1} + W'_{yf} y + b'_f) \\
i'_t = \sigma(W'_{xi} q_t + W'_{hi} h'_{t-1} + W'_{yi} y + b'_i) \\
o'_t = \sigma(W'_{xo} q_t + W'_{ho} h'_{t-1} + W'_{yo} y + b'_o) \\ 
g'_t= \phi  (W'_{xg} q_t + W'_{hg} h'_{t-1} + W'_{yg} y + b'_g) \\
c'_t = f'_t \odot c'_{t-1} + i'_t \odot g'_{t} \\
h'_t= o'_t \odot \phi( c'_t)
\end{array}
\end{equation}
where $W'_*$ and $b'_*$ are the parameters, which need to be learned. $y$ is the concatenated feature, mentioned in Eq. \ref{Eq.mean}. $h'_0  \in \mathbb{R}^{ D_{h'} \times 1}$ and $c'_0 \in  \mathbb{R}^{D_{h'} \times 1}$ are initialized vectors. Finally, we use a softmax layer to estimate the conditional probability distribution over $s_{t+1}$.
\begin{equation}
\begin{array}{l}
\label{euq.pro}
 P(s_{t+1}|s_{<t},\textbf{V}) = softmax(W_f h'_t + b_f) \ \ \ \ \ t \in \{1,...,N_s\} 
 \end{array}
\end{equation}
where  $W_f  \in \mathbb{R}^{ D_{s} \times D_{h'}}$ and $b_f \in \mathbb{R}^{ D_{s}}$ are the parameters. If the input is represented as $x\in \mathbb{R}^{ D_{s} \times 1}$, the softmax function can be expressed as:
\begin{equation}
\begin{array}{l}
\label{euq.softmax}
softmax(x_i) = \frac{e^{x_i}}{\sum_{j=1}^{D_s} e^{x_j}}\ \ \ \ \ i \in \{1,...,D_s\} 
 \end{array}
\end{equation}

\subsection{Loss Function}
Previous works \cite{CAP:MP-LSTM,CAP:S2VT,CAP:p_RNNYu_2016_CVPR} defined their loss functions based on maximum likelihood estimation (MLE). In this work, we follow them to define our loss function by optimizing the log-likelihood:

\begin{equation}
\begin{array}{l}
\label{euq.loss}
\mathcal{L} = \log P(\textbf{S}|\textbf{V}) \\
 =\sum_{t=1}^{N_s}  s^{T}_{t} \log P(s_{t}|s_{<t},\textbf{V})
 \end{array}
\end{equation}
By maximizing this loss function, we can estimate the parameters in the whole model. After extracting features with deep CNNs, we simultaneously train the rest of model (i.e. TP-LSTM, Mean Pooling Concatenating and Stack-LSTM in Fig. \ref{Fig.framework}). More specifically, we use back-propagation
through time (BPTT) algorithm to compute the gradients and
conduct the optimization with adadelta \cite{OPT:ADADELTA}.

\section{Experiments}

\label{sec:experimental}
We evaluate our model on the task of video captioning. We firstly study the performance of different features, and then evaluate the influence of different hyper-parameters. Finally, we compare our model with the state-of-the-art methods. 

\subsection{Datasets}
In our experiments, we use two public video captioning benchmarks that have been widely used in many other works.

\textbf{The Microsoft Video Description Corpus (MSVD).} This dataset is proposed by Chen \textit{et al.} \cite{Dataset:msvd}. There are $1,970$ short video clips collected from YouTube and about 80,000 descriptions collected by Amazon Mechanical Turkers (AMT) in this dataset, and an average length of each video clip about 9s. Each video clip has an average of forty descriptions. And this dataset is open-domain and covers a wide range of topics such as people, animals, sports, actions, music, scenarios, landscapes etc. In total, all the descriptions contain nearly $16,000$ unique vocabularies. Following previous work  \cite{CAP:EmbedPan2015Jointly,CAP:HRNNPan2016Hierarchical,CAP:p_RNNYu_2016_CVPR}, we split this dataset into a training, a validation and a testing dataset with $1200$ ($60$\%), $100$ ($5$\%) and $670$ ($35$\%) video clips, respectively.

\textbf{MSR Video to Text (MSR-VTT).}
 Xu \textit{et al.} \cite{Dataset:msr-vtt} collected this dataset by a commercial video search engining.
 It's a new large-scale and open-domain video captioning benchmark for supporting video understanding, especially for the task of automatically describing videos. There are  $10$K video clips and $200$K descriptions in this dataset, collected by  Amazon Mechanical Turkers workers (AMT) same as MSVD dataset, about 20 sentences for each short video. 
 It covers about 20 categories and diverse visual content. 
 The updated version contains many quality sentences, so we implement our experiments on the updated version.
 This dataset is divided into three subsets: $65$\% for training, $5$\% for validating and $30$\% for testing, corresponding to 6,513, 497 and 2,990 clips. 

\subsection{Evaluation Metrics}
Following previous works \cite{CAP:EmbedPan2015Jointly,CAP:S2VT,CAP:softattYao2015Describing}, for evaluating the performance of our method, we utilize the following three evaluation metrics: BLUE \cite{EVA:BLUE}, METEOR \cite{EVA:METEOR}, and CIDEr \cite{EVA:CIDEr}.

\subsection{Implementation Details }
\textbf{Preprocessing.} For preprocessing the descriptions of MSVD dataset, firstly we convert sentences to lower cases, and then use the wordpunct\_tokenizer in NLTK  \footnote{http://www.nltk.org/index.html} library to tokenize sentences and remove punctuations. Finally, we obtain a dictionary of 15,903 in size on the training splits. 

For preprocessing the descriptions of MSR-VTT dataset, we directly split descriptions with a blank space,  because they have been tokenized. As a result, we can obtain a dictionary of 23,662 in size on the training splits. In this experiment, we only take words which appear more than two times as the dictionary. Finally, we get a dictionary of 13,626 in size.

For the visual features, we use same method to extract features on both two datasets. For the spatial features, thanks to the ResNet-152 achieved the great results in image classification and video captioning \cite{CAP:EmbedPan2015Jointly,CAP:BAttri,CAP:S2VT}, we use a per-trained ResNet-152 on ImageNet \cite{Dataset:Imagenet} to extract visual features. At first, we select equally-spaced 30 frames from each video, then feed them into the per-trained ResNet-152 to extract features from the $pool5$ layer. Finally, we  get a $2048 \times 30$ feature matrix for each video. For the temporal features, inspired by \cite{VACT:two-streamfusion}, we first transform RGB images to optical flow images \cite{METH:Opt_flow} stacking with 10 frames, then we use a fine-tuned ResNet-152 \cite{VACT:two-streamfusion} on UCF101 to extract features from $pool5$ layer. As a result, we obtain a $2048 \times 30$ feature matrix. Next, we concatenate spatial and temporal features together and then feed them into our model. In our experiments, $D_v=4096$ and $N_v=30$.

\textbf{Training Details.} In the training phase, sentences in corpus are varying lengths, thus we add a begin-of-sentence flag $<$BOS$>$ to start each sentence and an end-of-sentence flag $<$EOS$>$ to end each sentence. In the testing phase, we input $<$BOS$>$ flag into our model to trigger the process of sentence generation.  Beam search method, a heuristic search algorithm based on greedy algorithm, is utilized to find a sentence, which has the max partial probability. In addition, the width of the beam search is set as 5.   

In addition, all the LSTM unit sizes are set as 512 ($D_h=D_q=D_{h'}=512$) and the word embedding size is set as 512 ($D_m=512$), empirically. In our experiments, we throw away the sentences whose length is more than 30, thus $N_s < 30$. The batch sizes are set as 64  on MSVD dataset and 256 on MSR-VTT dataset.  We apply the back-propagation through time (BPTT) algorithm to compute the gradients of the parameters and conduct the optimization with adadelta \cite{OPT:ADADELTA}. In addition, we set the learning rate as $10^{-4}$ to avoid the gradient explosion. We utilize dropout regularization with the rate of 0.5 in all layers and clip gradients element wise at 10. We stop training our model until 500 epoches are reached or until the evaluation metric does not improve the validation set at the patience of 20. Moreover, we utilize Theano \cite{Framewor:theano_1} framework to conduct our experiments.

All experiments are conducted on the Ubuntu 14.04 with Intel(R) Xeon(R) CPU E5-2650 v3 @ 2.30GHz and GeForce GTX TITAN X (Pascal) GPU.

\subsection{Experiments on MSVD}

For verifying the effectiveness of our framework, we design following experiments:

\textbf{Effectiveness of Different Features}. C3D features \cite{Net:C3DDuTran} are widely used for video captioning \cite{CAP:p_RNNYu_2016_CVPR,CAP:MFA_long,CAP:HRNNPan2016Hierarchical}.  In this experiment, we evaluate the influence of spatial ResNet-152 feature (res\_s) and compare our temporal ResNet-152 (res\_t) features with the C3D features. The baseline is our model without the TP-LSTM part. The experimental results are shown in Tab. \ref{Tab.feature}. From Tab. \ref{Tab.feature}, we can see that simply applying spatial ResNet-152 is quite effective for video captioning with B@4 (51.5\%), M (33.5\%) and C (75.8\%). Making use of both res\_s and c3d,  B@1, B@2, B@3 and B@4 improves, but M and C drops. In terms of video captioning evaluation, METEOR and CIDE are more reliable than BLEU.  Tab. \ref{Tab.feature} also shows that res\_s and rest\_t performs best in terms of B@4, M and C. Therefore, we prove that our res\_t performs better than c3d for video captioning. In the following experiments, all the models take both res\_s and res\_t.

\begin{table}[t]
	\centering
	\caption{Performances of our model with different features, where res\_s stands for the the spatial ResNet-152 feature, res\_t stands for the the temporal ResNet-152 feature, c3d stands for the C3D feature. B, M and C are short for BLUE, METEOR, and CIDEr. All values are reported as percentage (\%).}
		\label{Tab.feature}
	\begin{tabular}{c||c|c|c|c|c|c}
		\hline
		model&B@1&B@2&B@3&B@4&M&C\\
		\hline
baseline(res\_s)&81.7&71.1&61.9&51.5&33.5&75.8\\
baseline(res\_s+c3d)&\textbf{82.6}&\textbf{72.1}&\textbf{62.8}&52.1&32.2&63.6\\
baseline(res\_s+res\_t)&81.6&71.1&62.2&\textbf{52.7}&\textbf{34.3}&\textbf{75.9}\\
		\hline
	\end{tabular}
\end{table}

\textbf{The Effect of Segmentation Numbers}. In order to explore the effectiveness of temporal pooling, we study the influence of the segmentation numbers, represented as $N_e$. In this experiment, we set $N_e = 1$ (average whole features), $N_e = 3$ and  $N_e = 30$ (without average operation) and the experiments are shown in Tab. \ref{Tab.Ne}, where baseline stands for our approach without TP-LSTM. From Tab. \ref{Tab.Ne}, we can see that when $N_e = 3$, our model achieves better results than $N_e = 1$ and $N_e = 30$. When $Ne = 1$, the model ignores the temporal variance between long-range video shots.  When $Ne=30$, the model ignores the temporal invariance in a short video shot. This proves that reasonable number of segments can improve the performance of video captioning. Compared with the baseline, our model (TS-LSTM $N_e = 3$ ) achieves better results with 2.2\%, 2.7\%, 2.3\%, 1.8\% 0.2\% and 3.4\% increases on BLUE-1, BLUE-2, BLUE-3, BLUE-4, METEOR and CIDEr, respectively. Therefore, in the following experiments, we set $N_e = 3$.

\begin{table}[t]
	\centering
	\caption{Performances of our model with different $N_e$. All models except baseline use spatial ResNet-152 feature and temporal ResNet-152 feature. B, M and C are short for BLUE, METEOR, and CIDEr. All values are reported as percentage (\%).}
		\label{Tab.Ne}
	\begin{tabular}{c||c|c|c|c|c|c}
		\hline
		model&B@1&B@2&B@3&B@4&M&C\\
		\hline
baseline (res\_s+res\_t) &81.6&71.1&62.2&52.7&34.3&75.9\\
TS-LSTM($N_e=1$)(res\_s+res\_t)&83.0 &72.1 &62.6 &52.8 &33.7 &77.2 \\
TS-LSTM($N_e=30$)(res\_s+res\_t)&82.4 &72.1 &63.1 &53.3 &34.0 &76.7 \\
TS-LSTM($N_e=3$)(res\_s+res\_t)&\textbf{83.8} &\textbf{73.8} &\textbf{64.5} &\textbf{54.5} &\textbf{34.5} &\textbf{79.3} \\
		\hline
	\end{tabular}
\end{table}
\renewcommand\arraystretch{1}
\begin{table}[t]
	\centering
	\caption{\eat{(V), (G), (C), (3D) and (R) stands for the VGGNet, GoogLeNet, C3D, 3D-CNN, and ResNet, respectively.}  BLEU@N (B@N), METEOR (M), and CIDEr(C) scores of our model and other state-of-the-art methods. This experiment is conducted on the MSVD dataset. All values are reported as percentage (\%).}
	\begin{tabular}{c||c|c|c|c|c|c}
		\hline
		Model&B@1&B@2&B@3&B@4&M&C\\
		\hline
	    \hline
	    \eat{
		LSTM-E(V+C)\cite{CAP:EmbedPan2015Jointly}&78.8&66.0&55.4&45.3&31.0&-\\
		SA(G+3D)\cite{CAP:softattYao2015Describing}&80.0&64.7&52.6&42.2&29.6&51.7\\
		HRNE-AT(G+C)\cite{CAP:HRNNPan2016Hierarchical}&81.1&68.6&57.8&46.7&33.9&-\\
		h-RNN(V+C)\cite{CAP:p_RNNYu_2016_CVPR}&81.5&70.4&60.4&49.9&32.6&65.8\\
		MFA-LSTM(R+C)\cite{CAP:MFA_long}&82.9&72.0&62.7&52.8&33.4&68.9\\
		LSTM-TSA(V+C)\cite{CAP:Pan_2017_CVPR}&82.8 &72.0 &62.8 &52.8 &33.5 &74.0 \\
		hLSTMat(R)\cite{CAP:adpt_Song}&82.9 &72.2 &63.0 &53.0 &33.6 &73.8 \\
		\hline
		TP-LSTM(R\_s+R\_t)&\textbf{83.8} &\textbf{73.8} &\textbf{64.5} &\textbf{54.5} &\textbf{34.5} &\textbf{79.3} \\}
		MP-LSTM\cite{CAP:MP-LSTM}&-&-&-&33.3&29.1&-\\
		SA-LSTM\cite{CAP:softattYao2015Describing}&80.0&64.7&52.6&42.2&29.6&51.7\\
		LSTM-E\cite{CAP:EmbedPan2015Jointly}&78.8&66.0&55.4&45.3&31.0&-\\
		HRNE-AT\cite{CAP:HRNNPan2016Hierarchical}&81.1&68.6&57.8&46.7&33.9&-\\
		h-RNN\cite{CAP:p_RNNYu_2016_CVPR}&81.5&70.4&60.4&49.9&32.6&65.8\\
		M3-LSTM\cite{CAP:M3_Wang}&82.5 &72.4 &62.8 &52.8 &33.3 &- \\
		MFA-LSTM\cite{CAP:MFA_long}&82.9&72.0&62.7&52.8&33.4&68.9\\
		LSTM-TSA\cite{CAP:Pan_2017_CVPR}&82.8 &72.0 &62.8 &52.8 &33.5 &74.0 \\
		hLSTMat\cite{CAP:adpt_Song}&82.9 &72.2 &63.0 &53.0 &33.6 &73.8 \\
		\hline
		TS-LSTM&\textbf{83.8} &\textbf{73.8} &\textbf{64.5} &\textbf{54.5} &\textbf{34.5} &\textbf{79.3} \\		
		\hline
	\end{tabular}
	\label{Tab.comparebest}
\end{table}
\textbf{Comparing with existing methods}. To verify the availability of our model, we compare our results with the following methods:
\begin{itemize}
	\item MP-LSTM. Venugopalan \textit{et al.} \cite{CAP:MP-LSTM} used a mean-pooling layer to dispose all extracted frame-level features, then stacked two LSTM layers to explore semantic information.
	\item SA-LSTM. Yao \textit{et al.} \cite{CAP:softattYao2015Describing} introduced a temporal attention mechanism to automatically select the relevant frames, and combined with the spatial temporal 3-D convolutional
	neural network (3D-CNN) features, the model achieve the great results on the video captioning task. 
	\item LSTM-E. Pan \textit{et al,} \cite{CAP:EmbedPan2015Jointly} assumed that a low-dimensional  embedding exists for the representation of video and sentence, thus they mapped the video features and sentence features to the visual-semantic embedding and minimized the relevance loss to adequately explore the semantic information from videos.
	\item HRNE-AT. Pan \textit{et al.} \cite{CAP:HRNNPan2016Hierarchical} proposed a Hierarchical Recurrent Neural Encoder (HRNE) structure, which stacks a short LSTM on a long LSTM for adequately exploring the temporal information of a video. 
	\item h-RNN. Yu \textit{et al.} \cite{CAP:p_RNNYu_2016_CVPR} designed a sentence generator and a paragraph generator to generate paragraphs.
	The paragraph generator is stacked on the sentence generator and receives the state of the sentence generator, then initials the sentence generator.
	\item M3-LSTM. Wang \textit{et al.} \cite{CAP:M3_Wang} designed a visual and semantic shared memory structure for achieving the long-term visual-semantic dependency to further guide global visual attention. In this way, the model can learn an effective mapping from visual space to language space.
 	\item MFA-LSTM. Long \textit{et al.} \cite{CAP:MFA_long} selected the most frequent subject and verb across captions of each video, and took these as the semantic attributes and used a multi-modal attention mechanism to explore the semantic information from videos.
	\item LSTM-TSA. Pan \textit{et al.} \cite{CAP:Pan_2017_CVPR} introduced the Multiple Instance Learning (MIL). A weakly-supervised method was proposed to learn attribute detectors and  great results were achieved.
	\item hLSTMat. Song	\textit{et al.} \cite{CAP:adpt_Song} proposed a adjusted temporal attention mechanism, which can automatically decide whether to depend on the visual features or the semantic information, to improve the attention mechanism on video captioning.
\end{itemize}

\subsection{Comparison results on MSVD}
In this experiment, we firstly compare our method with the existing methods on the MSVD dataset and the results are shown in Tab. \ref{Tab.comparebest}. From Tab. \ref{Tab.comparebest}, we can see that our model obtains the best performance. In particular, the BLEU-4 of our model reaches 54.5\%, making a great improvement over h-RNN, MFA-LSTM, LSTM-TSA, hLSTMat by 4.6\%, 1.7\%, 1.7\%, 1.5\%, respectively. The METEOR of our model is 34.5\%, which outperforms h-RNN, MFA-LSTM, LSTM-TSA, hLSTMat by 1.9\%, 1.1\%, 1.0\%, 0.9\%, respectively.

\begin{figure*}
	\centering
	\includegraphics[width=1.0\textwidth,height=7cm]{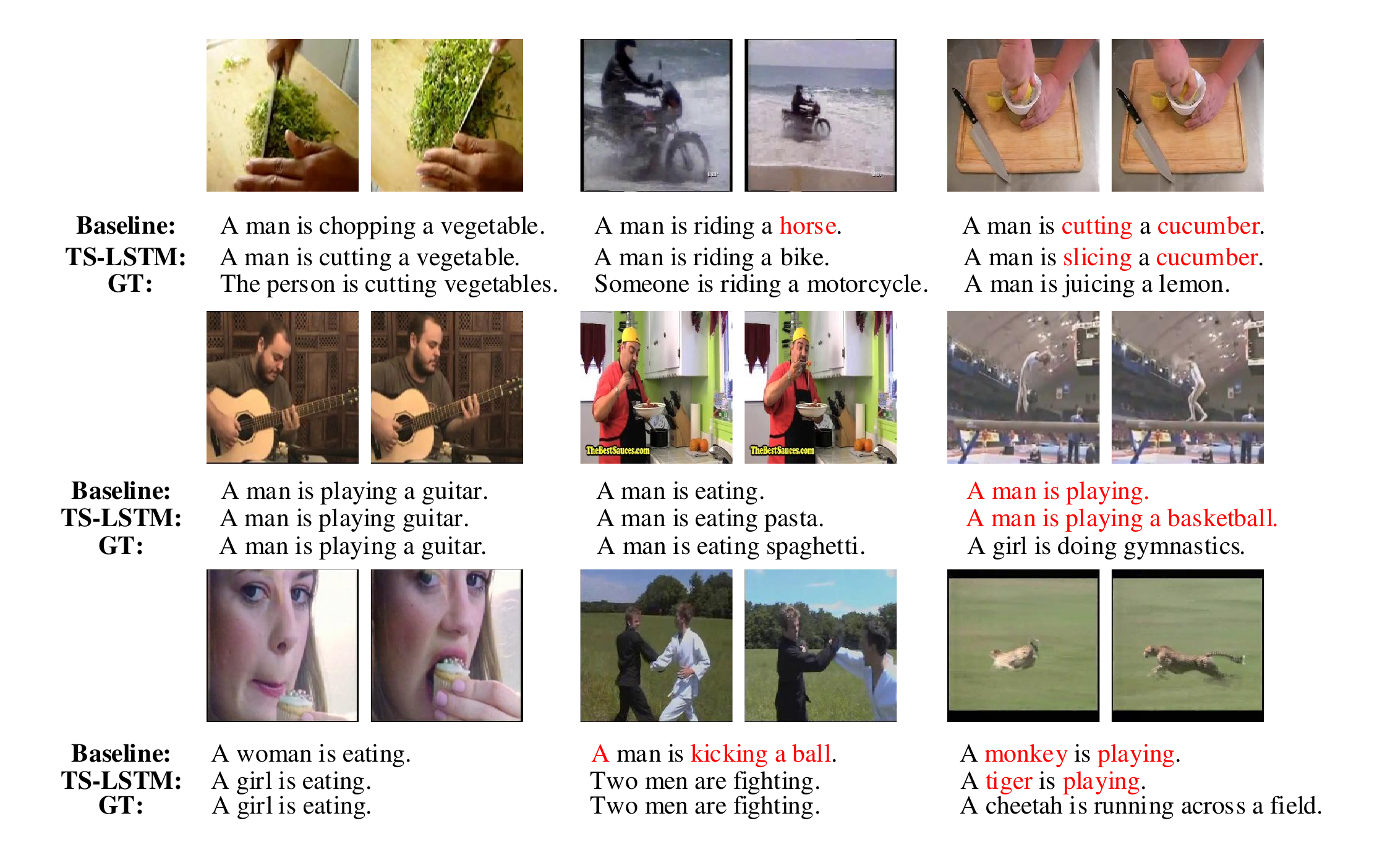}
	\caption{Some example sentences on the MSVD dataset. These sentences are generated by our model TS-LSTM and Baseline. GT denotes the ground truth. The imprecise words are labeled with red color.}
	\label{Fig.example}
\end{figure*}

In Fig. \ref{Fig.example}, we show some example sentences generated by our TS-LSTM model and our baseline mentioned in Section 4.4. The first column shows that both TS-LSTM and baseline can generate correct sentences to describe each video. From the second column, we have the following observations: 1) TS-LSTM model can generate sentences with accurate words to describe objects within a video, such as ``bike" in the top video. 2) Compared with baseline, TS-LSTM is able to provide more detailed information for describing video contents. For instance, in the middle video, TS-LSTM indicates that a man is ``eating pasta" instead of just ``eating". 3) For the bottom video in the second column, it shows that TS-LSTM has ability to calculate the number of objects within a video.  In addition, the third column introduces some wrong examples. For the bottom video in the third column, TS-LSTM and baseline generate two sentences: ``a monkey is playing" and ``a tiger is playing", respectively. Both of them are incorrect due to the following reason that the MSVD dataset contains few videos about ``cheetah". Therefore, the trained models both encounter an over-fitting problem.

\subsection{Comparison results on MSR-VTT}
To further illustrate the performance of our model, we compare our model with the state-of-the-art methods on the MSR-VTT dataset, which has the largest number of video-sentence pairs. The experimental results are shown in Tab.  \ref{Tab.comparebest_vtt}.

\begin{table}[t]
	\centering
	\caption{\eat{(V), (G), (C), (3D) and (R) stands for the VGGNet, GoogLeNet, C3D, 3D-CNN, and ResNet, respectively.}  BLEU@N (B@N), METEOR (M), and CIDEr(C) scores of our model and other state-of-the-art methods. This experiment is conducted on MSR-VTT dataset. All values are reported as percentage (\%).}
	\begin{tabular}{c||c|c|c|c|c|c}
		\hline
		Model&B@1&B@2&B@3&B@4&M&C\\
		\hline
	    \hline
		MP-LSTM\cite{CAP:MP-LSTM} &81.7&65.1&48.5&35.8&25.3&-\\
		SA-LSTM\cite{CAP:softattYao2015Describing} &\textbf{82.3}&\textbf{65.7}&49.7&36.6&25.9&-\\
		\eat{SA-LSTM\cite{CAP:softattYao2015Describing} &72.2&58.9&46.8&35.9&24.9&-\\}
		M3-LSTM\cite{CAP:M3_Wang}&73.6 &59.3 &48.3 &38.1 &26.6 &- \\
		hLSTMat\cite{CAP:adpt_Song} &-&-&-&38.3&26.3&-\\
		MFA-LSTM\cite{CAP:MFA_long} &-&-&-&39.2&26.6&\textbf{44.6}\\
		TS-LSTM&77.6&64.0&\textbf{51.3}&\textbf{39.9}&\textbf{27.1}&43.8\\
		\hline
	\end{tabular}
	
	\label{Tab.comparebest_vtt}
\end{table}
From Tab. \ref{Tab.comparebest_vtt}, we have the following observations. Firstly, our model achieves the best performance on BLEU-3 (51.3\%), BLEU-4 (39.9\%) and METEOR (27.1\%). Compared with MP-LSTM, SA-LSTM, M3-LSTM, hLSTMat, MFA-LSTM, our model improves the BLEU-4 by 4.1\%, 3.3\%, 1.8\%, 1.6\%, 0.7\%, respectively, and increases the METEOR by 1.8\%, 1.2\%, 0.5\%, 0.8\%, 0.5\%, respectively.


\section{Conclusion}
\label{rec:conclution}
In this paper, we present our temporal spatial LSTM network (TS-LSTM), a video level framework that aims to model long-term temporal dynamics and integrates dynamics with spatial and temporal features to improve video captioning. As demonstrated on two challenging datasets, this work outperforms the existing methods while keeping a reasonable computation cost. In this framework, the TP-LSTM is proposed to explore the long-range structure by taking segments of video visual and motion features as inputs, and produces informative long-term dynamics for video captioning. The experimental results show the effectiveness of our proposed approach.

\bibliographystyle{spmpsci}      
\bibliography{ref}   

\end{document}